\documentclass[conference]{IEEEtran}

\ifCLASSINFOpdf

\else
 
\fi

\usepackage{doc}
\usepackage{makeidx}
\usepackage{graphicx}
\usepackage{subfig}
\usepackage{float}

\begin{document}

\title{On Evaluating Power Loss with HATSGA Algorithm for Power Network Reconfiguration in the Smart Grid
}

\author{\IEEEauthorblockN{Flavio Galvao Calhau}
\IEEEauthorblockA{Federal University of Bahia\\
Email: flavio.calhau@ufba.br}
\and
\IEEEauthorblockN{Alysson Pezzutti\\}
\IEEEauthorblockA{Salvador University\\
Email: pezzuttinho@gmail.com}
\and
\IEEEauthorblockN{Joberto S. B. Martins}
\IEEEauthorblockA{Salvador University\\
Email: joberto.martins@unifacs.br}
}

\maketitle

\begin{abstract}
  This paper presents the power network reconfiguration algorithm HATSGA with a ``R" modeling approach and evaluates its behavior in computing new reconfiguration topologies for the power network in the Smart Grid context. The modelling of the power distribution network with the language ``R" is used to represent the network and support computation of distinct algorithm configurations towards the evaluation of new reconfiguration topologies. The HATSGA algorithm adopts hybrid Tabu Search and Genetic Algorithm strategy and can be configured in different ways to compute network reconfiguration solutions. The evaluation of power loss with HATSGA uses the IEEE 14-Bus topology as the power test scenario. The evaluation of reconfiguration topologies with minimum power loss with HATSGA indicates that an efficient solution can be reached with a feasible computational time. This suggests that HATSGA can be potentially used for computing reconfiguration network topologies and, beyond that, it can be used for autonomic self-healing management approaches where a feasible computational time is required.
  
  
\end{abstract}

\IEEEpeerreviewmaketitle

\section{Introduction}

Smart Grid represents a modern vision of a dynamic electricity grid, in which electricity and related information flow together in real time, allowing near-zero economic losses in the event of outages and power quality disturbances. All of this being supported by a new energy infrastructure built on top of communication channels, distributed intelligence and possibly clean power \cite{Sarvapali:2012}.

Outages on power supply are inevitable, whether for system components preventive maintenance interventions or operation of a protective device in the presence of a failure, among other possibilities. A maneuver plan must exist to deal with the effects resulting from interruptions, to minimize the affected area and restore power supply of such areas as soon as possible.

The computation of distribution network reconfiguration consists basically in finding, among all possible combinations, a set of configurations that (i) minimize the number of switching operations while keeping the radial structure of the network, (ii) reduce the energy loss and (iii) reduce the number of consumers without electricity supply. In addition, it is necessary to safeguard voltage levels restrictions at the point of load, maximize current flow in the branches and safeguard lines flow capacity and nominal power of the transformers.

In brief, this work aims to evaluate HATSGA algorithm behavior for the problem of distribution power network reconfiguration.

\section{RELATED WORK}

In recent years, significant research has been done to minimize power loss in the process of reconfiguring distribution power systems.

The reconfiguring of distribution power system has a combinatorial nature \cite{Carreno} and the search of the best computational time for supporting the decision-making process in real time has been focused on recent research \cite{Thomas}, \cite{Haughton}, \cite{Merdan}.

Tabu search algorithm has been used for several combinatorial optimization solutions \cite{Ramon}, \cite{R.Cherkaoui}.  In \cite{Thakur} Tabu search is shown as a meta-heuristic method for network reconfiguration problem in radial distribution systems. In \cite{abdelaziz2010distribution} proposed a method of reconfiguring networks through a Tabu search algorithm which was modified focusing on reducing the opening and closing keys and minimizing the loss of active system. The proposed HATSGA algorithm enhanced the Tabu list to have more data-related computation and reduce the computation time avoiding exhaustive computation.

In \cite{Kng.Yi} a method based on genetic algorithms (GAs) is used to minimize loss in distribution systems by using encoding/decoding vertices to determine network configuration. The number based on the vertices was used in GAs for encoding/decoding of chromosomes and the system topology is radial. In general, using methods based on AI (Artificial Intelligence) proved to be valuable in a variety of applications \cite{KIRAN:2013}. HATSGA does use a distinct genetic algorithmic approach by using “elitism” to choose among potential solutions. This is effectively different from using AI techniques to compute the entire reconfiguration solution.

Methods based on hybrid algorithms are indicated as having the potential to significantly reduce computational time of solutions \cite{KIRAN:2013}.

\section{MODELING THE IEEE 14-BUS WITH ``R" FOR POWER LOSS COMPUTATION}

Graphs are used to model the structure of electrical grids for various purposes, and to verify its resilience to outages. In this work the distribution network being reconfigured is modeled using graphs theory with language ``R".

Language ``R" was adopted due to its capability to semantically represent with flexibility the basic components of a distribution network (switches, branches, feeders, operational parameters, among others). In addition to that,``R" does present a series of programming flexibility aspects such as powerful data types representation and manipulation together with an attractive set of tools that are necessary for graph computation \cite{calhau_hybrid_2015}, \cite{FGCADVANCE:2014}.

\subsection {Generic modelling a distribution network with graphs and ``R"}

\begin{figure}[htb]
\centering
\includegraphics[width=0.45\textwidth]{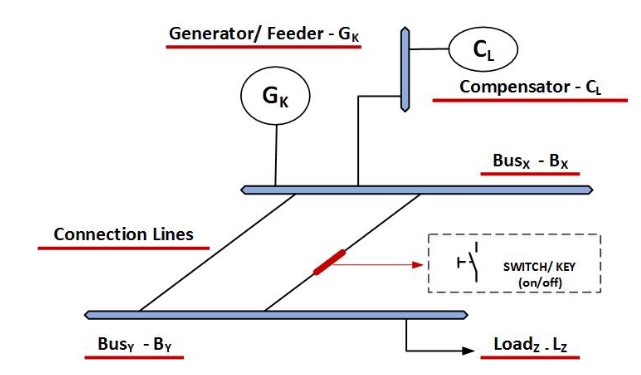}
\caption{Power distribution network elements.}
\label{fig:PowerDistributionNetworkElements}
\end{figure}

A power distribution network is composed by a set of elements where the main element is the “bus”. The buses are interconnected by connection lines.  Generators, loads and synchronous compensators are also attached to the buses in the network defining their characteristics as slack, load bus or generator bus (Figure \ref{fig:PowerDistributionNetworkElements}).

In the distribution network the connection lines may include switches. These switches can be on (closed - tie switch) or off (open – sectionalizing switch) to either have an effective connection through the line between buses or not.

\begin{figure}[H]
\centering
\includegraphics[width=0.5\textwidth]{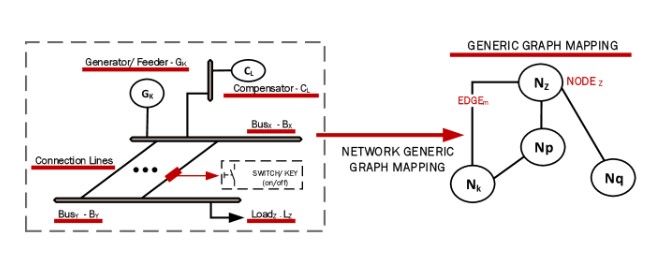}
\caption{Network Generic Graph Mapping.}
\label{fig:NetworkGenericGraphMapping}
\end{figure}

In terms of the graph theory, a power distribution network is generically mapped as a connected graph $G = \{N_z, E_m\}$ with a set of nodes $N$, $N = \{N_1, N_2, ... N_z\}$, and a set of edges $E$, $E = \{E_1, E_2, ... E_m\}$ (Figure \ref{fig:NetworkGenericGraphMapping}). Nodes in the graph (vertex) have degree $ND_i \geq 1$ and the network has a topology that may include loops or be radial with no loops. Edges interconnect nodes in the graph mapped topology - ($E_i = \{V_i, V_i\} \notin G)$.

The modeling of the power distribution network with language ``R" basically takes the nodes and edges used in the generic graph topology representing the network and adds a set of more specific mappings and powerful parameter representation as follows (Figure 3):


   \begin{figure}[ht]
     \subfloat{%
       \includegraphics[width=0.45\textwidth]{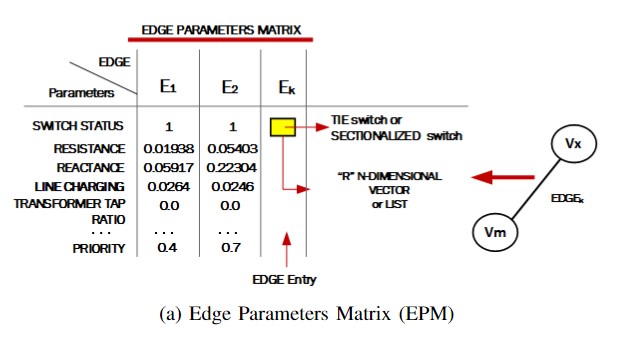}
     }
     \hfill
     \subfloat{%
       
       \includegraphics[width=0.45\textwidth]{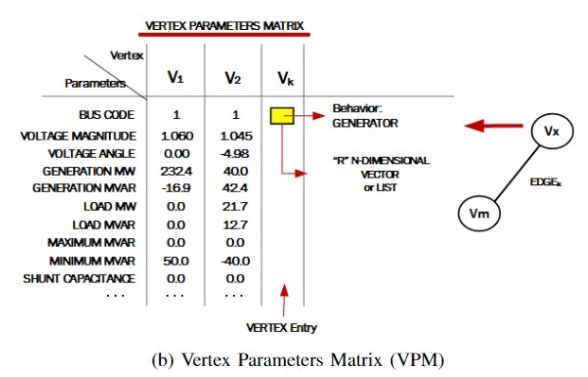}
     }
     \caption{Mappings parameter representation}
     \label{fig:MappingsParameterRprepresentation}
   \end{figure}

\begin{itemize}

		\item[--] Connection lines (network) are mapped to ``edges" (as in graph representation);
		\item[--] For each ``edge", it is defined using ``R" an entry (value) at the ``Edge Parameters Matrix – EPM" (Figure \ref{fig:MappingsParameterRprepresentation});
		\item[--] Connection lines with either tie switch (switch closed) or sectionalizing switch (switch open) behaviors are modeled as distinct type of edges (connecting or not vertexes); and
		\item[--] Vertex degrees, as in graph representation, are modeled with ``n" distinct edges between vertexes and associated EPMs entries.
		\item[--] Buses (network) are mapped to vertex;
		\item[--] For each vertice, it is defined using ``R" an entry (value) at the ``Vertex Parameters Matrix - VPM" (Figure \ref{fig:MappingsParameterRprepresentation});
		\item[--]Slack bus, load bus and generator bus (modeling generators/feeders) behaviors are modeled as different types of vertex using VPM;
		
\end{itemize}

The Vertex Parameters Matrix (VPM) (Figure \ref{fig:MappingsParameterRprepresentation}) shows that data format for input the parameters at each bus in the power system. Thus, at all the buses, the Voltage magnitude and angle, the loads (P-Load \& Q-Load), generation (P-Gen \& Q-Gen), reactive power limits (Q-min \& Q-max), and Shunt Capacitance should be specified (usually set to zero when not needed).

VPM is composed by a set of N-dimensional vectors or lists using ``R" data types representation. Each entry on the matrix (vector or list) stores all parameters or attributes associated with the vertex modeled as a specific bus (including its behavior as slack, generator or load bus). Parameters values, attributes, properties and parameter descriptions represented in the matrix include fundamentally all electrical parameters relevant (or not) for the computation. In addition, a set of user defined parameters is included in the matrix. In effect, the adopted approach supported by ``R" modelling allows a rather general description of the bus parameters and, beyond that, can be easily adjusted or updated by including new elements. For ``R" computation, any additional adjustment in the matrix modelling corresponds to redefining the dimension of the matrix and vectors used together with the description of the parameters.

The Edge Parameters Matrix (EPM) (Figure \ref{fig:MappingsParameterRprepresentation}) stores and represents transmission line parameters and data to be used by HATSGA. Parameters like R (resistance), X (reactance) and Line Charging (1/2B) are electrical properties of the line, while the tap ratio represents the transformer tap settings and line limit represents the thermal rating of the line.

The Edge Parameters Matrix (EPM) is equivalent to VPM and includes all edge parameters, attributes and properties relevant to the computation of network reconfiguration.  In terms of the ``R" computation it is important to mention that each individual element of the matrix entry can, in turn, have a multi-dimensional representation allowing very powerful representation of the modeled network. In addition to that, various data types may be used when modelling the network with VPM and EPM and on-the-fly modifications are allowed facilitating the computation.

\subsection {Modeling the IEEE 14-bus with ``R"}

\begin{figure}[]
\centering
\includegraphics[width=0.6\textwidth]{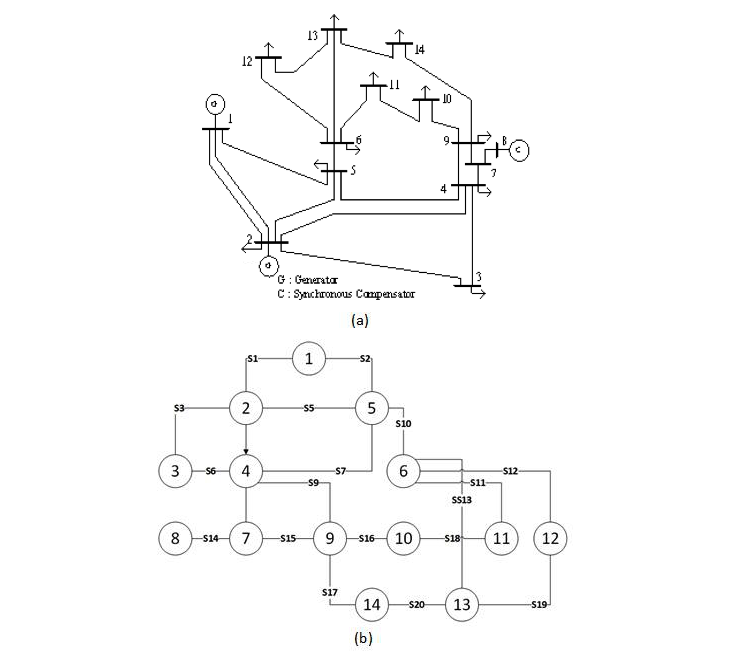}
\caption{(a) IEEE-14 bus test system, (b) Computational mapping model (graph) of IEEE-14 bus.}
\label{fig:ComputacionalModel}
\end{figure}

Figure \ref{fig:ComputacionalModel}(a) depicts the IEEE 14 bus system with 14 bus bars, and 20 links connecting them that will be used in the HATSGA simulation. Figure \ref{fig:ComputacionalModel}(b) is the corresponding graph mapped from the original IEEE 14 bus system.

The HATSGA algorithm uses in this implementation the minimum power loss as the objective function but allows the use of other general objectives like ``priority" (previewed in the EPM matrix) in addition to the minimum loss. This, for instance, can be helpful in cases where we want to reconfigure a network keeping certain areas with some priorities.

\section {HATSGA ALGORITHM}

The hybrid algorithm HATSGA computes a reconfiguration topology for a radial power distribution network modeled in ``R" language having as its basic constraint to minimize the power losses. In terms of the calculation, HATSGA algorithm adopted the Newton-Rapson method \cite{BENISRAEL} that, in turn, is based on the minimum power loss equation illustrated in equation \ref{eq:minimumlossmathmodel}. The values of the parameters of the matrices EPM and VPM (Figure \ref{fig:MappingsParameterRprepresentation}) are used.

\begin{equation} 
\label{eq:minimumlossmathmodel}
min f = \sum_{i=1}^{n} k_{i}r_{i}\frac{P^{2}_{i} + Q^{2}_{i}}{V^{2}_{i}}
\end{equation}

Where

\begin{itemize}
		\item n represents total numbers of branches (edges).
		\item r\textsubscript{i} - resistance of edge i.
		\item Q\textsubscript{i} - reactive power of edge i.
		\item P\textsubscript{i} - active power of edge i.
		\item V\textsubscript{i} - voltage on head node of branch i.
		\item k\textsubscript{i} - switch on branch i, 1 can equal to closed, 0 can equal to open.
\end{itemize}

 HATSGA's strategy minimizes the search space solution by eliminating topology configurations that do not comply with constraints criteria like radial configuration, resistance and reactance, among others.
 
In terms of the genetic algorithm strategy, HATSGA uses ``elitism", a technique inspired by evolutionary biology and natural selection. In this case, elitism is used to select potential topologies that can be used to compute new reconfiguration solutions by selecting the best or most adequate electrical parameter(s) that may lead to the best possible result in the computation of new minimal solutions for power loss. 

The Tabu Search algorithmic approach within HATSGA uses a list to store found solutions that should not be considered in the computation (forbidden).  That is used to do not visit solutions that were previously considered (illegal movements). Conventional Tabu List implementations consider all topologies but only stores the minimal power loss for the best solution found while computing. The Tabu List conventional implementations also uses a temporary list, meaning that after computation the Tabu list is deleted.

HATSGA uses a modified Tabu Search algorithm by introducing to the conventional ``tabu list" a set of associated parameters. In effect, the HATSGA Tabu List is a two-dimensional matrix composed by a list of open switches and the power loss computed for all the resulting topologies. Another aspect differentiating the conventional tabu list from the HATSGA's one is that the list is always kept during computation to allow some optimizations in terms of the computational time.

Distribution systems commonly operate with a radial topology. HATSGA imposes radiality constraint to compute solutions,  and the summary of the strategic phases of the algorithm are as follows:

\begin{itemize}

		\item The algorithm always starts with a radial topology obtained by the PRIM's algorithms that creates a spanning tree for the network topology;
		\item From the initial trigger topology, it is calculated the minimum power loss and the HATSGA Tabu List is initialized with its first entry (topology - set of open switches - power loss);
		\item HATSGA now takes the list of open switches (available at the modified Tabu List) and starts closing the switches creating, as such, sectorial loops (as a resulting property of the spanning tree algorithm and topology);
		\item For each sectorial loop found, HATSGA finds a new set of open switches that generates new sector set of radial topologies;
		\item For each sectorial loop found and set of derived radial topologies, HATSGA uses the “elitism” method to select the set of topologies for computing the power loss;
		\item This approach represents in effect a new embedded heuristic that effectively minimizes the set of possible solutions and reduces the overall computation time for the minimum power loss calculation;
		\item For each sectorial loop and topologies chosen by elitism the power loss is computed and the HATSGA modified Tabu List is updated with eventual new minimum power loss values;
		\item The process continues with the next open switches creating new loop sectors that are minimized in terms of topologies using elitism and so on till the entire list of open switches if sequentially considered.
		
\end{itemize}

\section {HATSGA Operational Model}
 
The HATSGA algorithm starts with a graph topology representing the power network (IEEE 14-bus in this example) supported by the ``R" language mappings (VPM, EPM matrixes and other relevant aspects of the mapping) as described in section III (Figure \ref{fig:NetworkGenericGraphMapping}).

\begin{figure}[]
\centering
\includegraphics[width=0.5\textwidth]{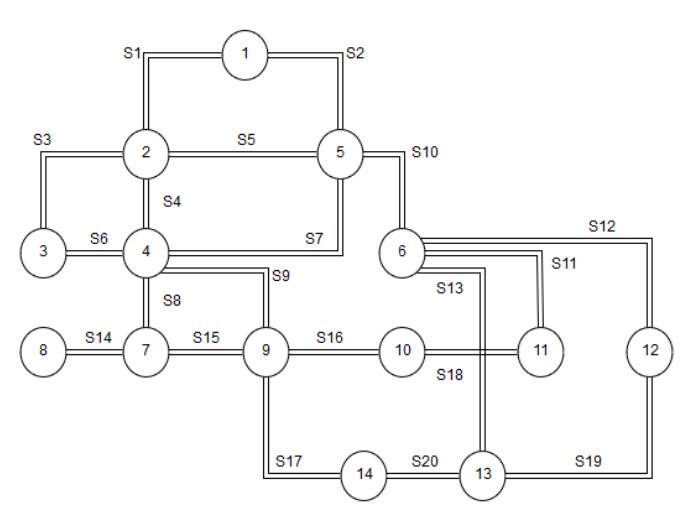}
\caption{The IEEE-14 bus test system graph map (nodes) with language ``R" support.}
\label{fig:NetworkGenericGraphMapping}
\end{figure}

\begin{figure}[H]
\centering
\includegraphics[width=0.5\textwidth]{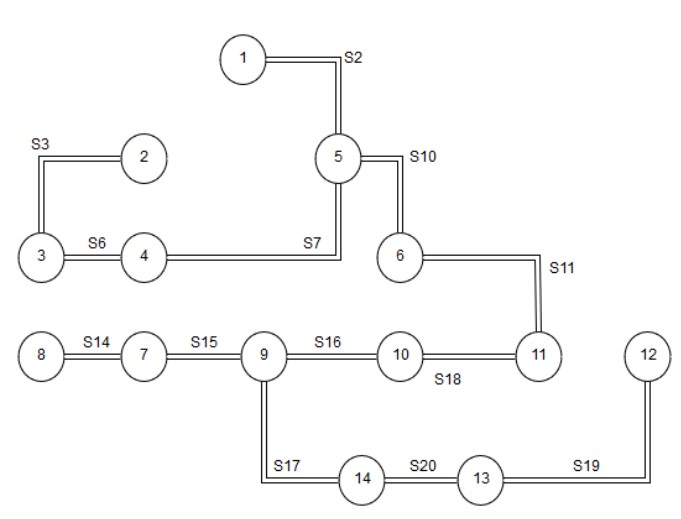}
\caption{An example of radial topology obtained with Prim algorithm (initial trigger topology).}
\label{fig:ExampleRadialTopologyPrim}
\end{figure}

HATSGA basic requirement is to start with a radial topology. Figure \ref{fig:ExampleRadialTopologyPrim} illustrates a possible radial topology (the initial trigger topology) obtained from the IEEE 14-bus graph with R mapping by using the Prim's algorithm. The Prim algorithm generates a minimal tree topology in which the sum of all edges is minimized keeping all vertices interconnected without loops.

The initial trigger topology has an associated list of open switches (Table \ref{openedswitches}) that will drive the remaining execution steps of the HATSGA algorithm.

\begin{table}[H]
\centering
\caption{Open Switches lists obtained with the Initial Trigger Topology}
\label{openedswitches}
\begin{tabular}{|l|l|}
\hline
\multicolumn{2}{|l|}{\textbf{Initial Trigger Topology (from IEEE-14bus)}} \\ \hline
Open Switches                & S1; S4; S5; S8; S9; S12; S13               \\ \hline
\end{tabular}
\end{table}

The minimal loss topology search realized by HATSGA uses a ``sector-based'' graph approach with an associated heuristic. The sector-based approach is effectively a partitioning strategy adopted in this graph-based computation that, in turn, makes use of the basic properties of the spanning-tree algorithms. The heuristic used for each sector computation will contribute to minimize the total computation time required to obtain the results and, as such, contributes to a more scalable algorithm execution.

\begin{figure}[H]
\centering
\includegraphics[width=0.4\textwidth]{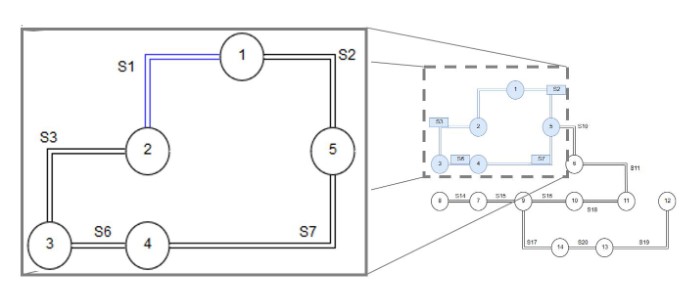}
\caption{Sector loop generated by closing S1 switch.}
\label{fig:sectorloop01}
\end{figure}

The HATSGA algorithm uses the list of open switches obtained from the initial trigger topology (Table \ref{openedswitches}) to create sectors with loops. Figure \ref{fig:sectorloop01} illustrates switch S1 being closed and, as such, creating a sector loop in the previously loop-free topology.

\begin{figure}[H]
\centering
\includegraphics[width=0.25\textwidth]{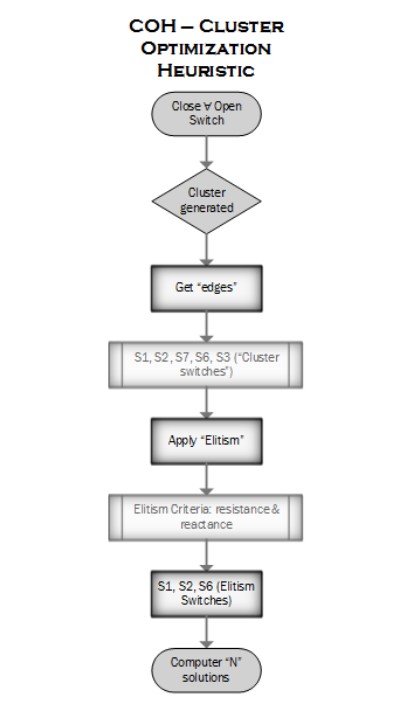}
\caption{The Cluster Optimization Heuristic (COH).}
\label{fig:COH}
\end{figure}

Once the loop is created, HATSGA gets the list of edges belonging to this sector loop (cluster) and, following that, will compute new reconfiguration topologies for this cluster applying a new heuristic (COH - Cluster Optimization Heuristic) based on ``elitism'' (Figure \ref{fig:COH}). In ``elitism'' at least one copy of the best solution is passed onto the new population (possibilities of solutions) so that the best solution can survive successive generations.

Table \ref{eliminatingloops} exemplifies the possible edges that might be used to eliminate loops (LoopList) in the graph and the edges selected (filtered) by the elitism heuristics (LoopElitism).

\begin{table}[H]
\centering
\caption{Possibilities of eliminating loops (with and without elitism) by manipulating the status of edges.}
\label{eliminatingloops}
\begin{tabular}{|l|c|}
\hline
\multicolumn{2}{|c|}{\textbf{\begin{tabular}[c]{@{}c@{}}Loop Switches \\ (cluster example in Figure \ref{fig:sectorloop01})\end{tabular}}} \\ \hline
\textit{LoopList}     & S1; S2; S3; S6; S7                                        \\ \hline
\textit{LoopElitism}     & S1; S6; S7                                                \\ \hline
\end{tabular}
\end{table}

\begin{figure}[H]
\centering
\includegraphics[width=0.45\textwidth]{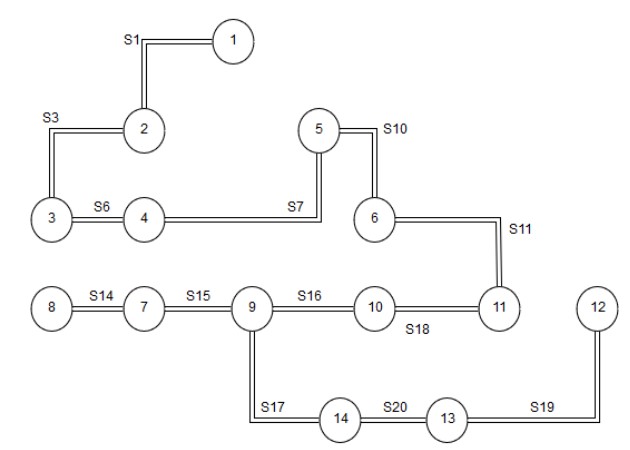}
\caption{Example of graph resulting from loop elimination computation generated in Figure \ref{fig:sectorloop01}.}
\label{fig:graphnoloopelitism01}
\end{figure}

Based on the ``LoopElitism'' switches (edges) list the HATSGA algorithm computes the minimal power loos for this cluster. Figure \ref{fig:graphnoloopelitism01} illustrates a topology obtained after computing the minimal power loss. 

HATSGA repeats this cluster-based computation process for the list of open switches obtained from the initial topology and selects the best minimal loss power computation among all this sector-based and elitist selected options.

\section {HATSGA Test Scenarios and Results}

The objectives of the HATSGA test scenarios are:

\begin{itemize}

		\item Evaluate HATSGA behavior regard minimum power loss computation results for distinct initial topologies generated by the PRIM's algorithm with different edge ``priorities'';
		\item To evaluate the computational time required to compute the solutions.
		
\end{itemize}

\subsection {HATSGA Test Scenarios}

HATSGA test scenarios uses the IEEE-14 test system, modeled on graph with 14 vertexes and 20 edges (Figure \ref{fig:NetworkGenericGraphMapping}) as the power system.

\begin{itemize}
	\item HATSGA evaluated metrics:
    \begin{itemize}
		\item System power loss;
		\item o	HATSGA system power loss computational time.
	\end{itemize}
	
	\item HATSGA System Parameters:
    \begin{itemize}
		\item Initial radial topology algorithm - Prim;
		\item Heuristic algorithm for Search Solution - Tabu Search and Elitism;
		\item Power flow method - Newton-Raphson;
		\item Priority (used effectively as edge weight in the computed Prim topology).
	\end{itemize}
	
	\item Power Distribution Workload Parameters:
    \begin{itemize}
		\item EPM Matrix;
		\item VPM Matrix.
	\end{itemize}
	
\end{itemize}

The results were obtained on a Macbook with an Intel Core i7 2.6 Ghz CPU and 8 GB RAM.

The power loss computation time was measured by the function (\textit{proc.time}) available with the ``R" programming environment. This function determines how much computational time (in seconds) the currently running R process (HATSGA code) has already taken. The \textit{proc.time} function essentially works as a stop-watch: you initialize it to a starting time, run all the desired code, and then stop it by subtracting the starting time from the ending time.

The overall expected result of the HATSGA algorithm is that it minimizes computation and keep at the same time the quality of the solution by using elitism with a set of electrical parameters that, potentially, lead to a minimum power loss.

Table \ref{hatsgaresults} shows the computed power system loss and computational time (in seconds) for 50 simulation runs where all initial trigger topologies are guaranteed to be distinct. The priority used (weight) for the initial topology edges are generated randomly with values from 1 to 1000. 

In terms of the computational time required to compute the power loss, the mean value obtained for IEEE 14-bus was 1.99 seconds with 95\% of confidence interval.

In terms of the power loss, the minimum value obtained was 4,25 with a mean computed value of 5.07.

In terms of the power loss computation, HATSGA gets similar results (minimal loss: 4,25; mean: 5,07; 95\% confidence interval) for distinct initial trigger topologies. In terms of the numbers involved in the simulation runs, there are for IEEE 14-bus 3909 possible of minimum spanning tree topologies and HATSGA, using its heuristics (Prim, Tabu and Genetic) reduces the number of generation trees to a number around 50 effectively possible and considered topologies.

In terms of the minimal power loss computed or, in other words, the quality of the solution, HATSGA algorithm behaves better than most algorithms found and documented in the literature obtaining the second-best quality result (Table \ref{tab:ResultsIEEE14Bus} \cite{NACHIMUTHU}). To the extent of our knowledge, no further detailed comparison can be made in relation to these referred solutions because there is no explicit information how they compute their results. Being so, the comparison is only in terms of the result obtained.

\begin{table}[H]
\centering
\caption{Hatsga simulation results}
\label{hatsgaresults}
\begin{tabular}{|c|c|c|c|}
\hline
\textbf{\begin{tabular}[c]{@{}c@{}}Simulation \\ Run\end{tabular}} & \textbf{\begin{tabular}[c]{@{}c@{}}Opened \\ Switch(es)\end{tabular}} & \textbf{\begin{tabular}[c]{@{}c@{}}System \\ Loss\end{tabular}} & \textbf{\begin{tabular}[c]{@{}c@{}}Computational \\ Time\end{tabular}} \\ \hline
1                                                                  & 5-1-3-9-7-13-17                                                       & 42.506                                                          & 2.14                                                                   \\ \hline
2                                                                  & 1-3-4-7-13-15-17                                                      & 56.688                                                          & 1.991                                                                  \\ \hline
3                                                                  & 5-7-6-15-1-13-17                                                      & 57.801                                                          & 2.084                                                                  \\ \hline
4                                                                  & 6-7-1-8-4-13-17                                                       & 42.506                                                          & 1.921                                                                  \\ \hline
5                                                                  & 1-3-4-7-17-13-15                                                      & 42.506                                                          & 1.987                                                                  \\ \hline
6                                                                  & 5-7-6-9-1-13-17                                                       & 42.506                                                          & 1.958                                                                  \\ \hline
7                                                                  & 1-3-4-9-7-13-17                                                       & 42.506                                                          & 2.002                                                                  \\ \hline
8                                                                  & 1-3-4-13-15-7-17                                                      & 62.429                                                          & 1.936                                                                  \\ \hline
9                                                                  & 7-1-3-9-4-13-17                                                       & 57.801                                                          & 2.013                                                                  \\ \hline
10                                                                 & 1-3-7-4-9-13-17                                                       & 42.506                                                          & 2.239                                                                  \\ \hline
11                                                                 & 1-3-4-7-13-15-17                                                      & 42.506                                                          & 1.958                                                                  \\ \hline
12                                                                 & 7-5-6-1-9-13-17                                                       & 42.506                                                          & 1.991                                                                  \\ \hline
13                                                                 & 1-3-7-9-4-13-17                                                       & 42.506                                                          & 1.959                                                                  \\ \hline
14                                                                 & 6-7-1-9-4-13-12                                                       & 42.506                                                          & 1.868                                                                  \\ \hline
15                                                                 & 1-4-5-7-13-15-17                                                      & 42.506                                                          & 1.997                                                                  \\ \hline
16                                                                 & 1-3-7-8-4-13-17                                                       & 57.801                                                          & 2.048                                                                  \\ \hline
17                                                                 & 1-3-7-9-4-17-13                                                       & 42.506                                                          & 2.292                                                                  \\ \hline
18                                                                 & 1-3-7-9-4-17-13                                                       & 42.506                                                          & 1.955                                                                  \\ \hline
19                                                                 & 2-3-4-9-7-17-13                                                       & 56.688                                                          & 1.921                                                                  \\ \hline
20                                                                 & 2-3-4-8-13-7-17                                                       & 42.506                                                          & 2.011                                                                  \\ \hline
21                                                                 & 6-7-1-9-4-13-12                                                       & 69.753                                                          & 1.965                                                                  \\ \hline
22                                                                 & 1-3-4-7-9-13-17                                                       & 42.506                                                          & 1.932                                                                  \\ \hline
23                                                                 & 5-7-6-15-13-1-17                                                      & 56.688                                                          & 1.891                                                                  \\ \hline
24                                                                 & 6-7-1-9-4-13-12                                                       & 56.688                                                          & 2.075                                                                  \\ \hline
25                                                                 & 2-3-4-7-9-13-17                                                       & 50.252                                                          & 2.061                                                                  \\ \hline
26                                                                 & 6-1-7-8-4-13-17                                                       & 42.506                                                          & 2.088                                                                  \\ \hline
27                                                                 & 1-3-4-7-13-15-17                                                      & 42.506                                                          & 1.914                                                                  \\ \hline
28                                                                 & 1-3-7-9-13-4-17                                                       & 42.506                                                          & 1.943                                                                  \\ \hline
29                                                                 & 1-3-7-9-4-13-17                                                       & 69.753                                                          & 2.011                                                                  \\ \hline
30                                                                 & 1-3-4-7-9-13-17                                                       & 42.506                                                          & 1.906                                                                  \\ \hline
31                                                                 & 1-3-4-8-7-13-17                                                       & 42.506                                                          & 2.059                                                                  \\ \hline
32                                                                 & 6-7-5-1-9-13-17                                                       & 42.506                                                          & 1.858                                                                  \\ \hline
33                                                                 & 1-3-7-8-13-4-17                                                       & 42.506                                                          & 1.93                                                                   \\ \hline
34                                                                 & 1-3-4-7-15-13-17                                                      & 42.506                                                          & 2.005                                                                  \\ \hline
35                                                                 & 1-3-4-7-13-15-17                                                      & 56.688                                                          & 2.075                                                                  \\ \hline
36                                                                 & 1-3-4-7-17-13-15                                                      & 50.252                                                          & 2.021                                                                  \\ \hline
37                                                                 & 1-3-4-7-8-13-17                                                       & 42.506                                                          & 2.311                                                                  \\ \hline
38                                                                 & 1-3-4-7-17-13-15                                                      & 69.753                                                          & 2.059                                                                  \\ \hline
39                                                                 & 1-3-4-8-7-13-17                                                       & 69.753                                                          & 1.884                                                                  \\ \hline
40                                                                 & 1-3-7-8-13-4-17                                                       & 42.506                                                          & 2.016                                                                  \\ \hline
41                                                                 & 1-3-4-7-8-13-17                                                       & 69.753                                                          & 2.048                                                                  \\ \hline
42                                                                 & 5-7-6-1-9-13-17                                                       & 42.506                                                          & 1.957                                                                  \\ \hline
43                                                                 & 2-3-4-7-9-13-17                                                       & 56.688                                                          & 2.006                                                                  \\ \hline
44                                                                 & 1-4-3-7-13-15-17                                                      & 42.506                                                          & 1.98                                                                   \\ \hline
45                                                                 & 1-4-5-7-13-15-17                                                      & 56.688                                                          & 1.843                                                                  \\ \hline
46                                                                 & 2-3-4-7-9-13-17                                                       & 57.801                                                          & 1.876                                                                  \\ \hline
47                                                                 & 1-4-5-7-9-17-13                                                       & 42.506                                                          & 1.918                                                                  \\ \hline
48                                                                 & 6-1-7-8-13-4-17                                                       & 42.506                                                          & 1.884                                                                  \\ \hline
49                                                                 & 1-3-7-9-13-4-17                                                       & 70.828                                                          & 1.769                                                                  \\ \hline
50                                                                 & 1-3-4-7-13-15-17                                                      & 42.506                                                          & 2.013                                                                  \\ \hline
\multicolumn{2}{|c|}{\textbf{Mean}}                                                                                                        & \textbf{5.068}                                                           & \textbf{1.991}                                                                  \\ \hline
\multicolumn{2}{|c|}{\textbf{Standard Deviation}}                                                                                          & \textbf{1.038 }                                                          & \textbf{0.104 }                                                                 \\ \hline
\multicolumn{2}{|c|}{\textbf{Confidence Interval 95\%}}                                                                                         & \textbf{{[}4.761 - 5.374{]}}                                             & \textbf{{[}1.961 - 2.022{]}}                                                    \\ \hline
\end{tabular}
\end{table}

\begin{table}[htb]
\centering
	\begin{centering}
		\caption{Power Loss computed with other algorithms.}
		\label{tab:ResultsIEEE14Bus}
		\begin{tabular}{lc}
		\hline
		Algorithm & \begin{tabular}{cc} Power Loss (MW)\end{tabular} \\
		\hline
		Original Network (IEEE-14) 		& 13.436 	\\
		PSO Algorithm 				&  9.9159 	\\
		MPSO Algorithm  				&  8.5053 	\\
		ABC Algorithm 				&  6.4611 	\\
		FSS Algorithm 				&  7.8457 	\\
		\textbf{HATSGA}   			&  \textbf{4.2796} 	\\
		GSA Algorithm 				&  3.2764 	\\
		\hline
		\end{tabular}
	\end{centering}
\end{table}

\section {Conclusion}

The hybrid algorithm (HATSGA) models the distribution network using the language ``R".  The ``R" modelling allowed a simple and efficient way to represent and compute minimal spanning tree algorithms with the radial topologic solutions required for the reconfiguration problem. ``R" language also allows the introduction of some semantics for network graphs representation that are essential for computing more elaborate, complex and qualified topologies in which various user requirements are considered.

HATSGA simulation results obtained for the IEEE 14-bus topology show a feasible response time of 1.9 seconds on a typical top-desk computer. That suggest firstly, that this algorithm can be used for computing reconfiguration network topologies nearly on-the-fly, at least, for networks as complex as IEEE 14-bus. Secondly, this suggests that HATSGA be used for autonomic self-healing management approaches in the Smart Grid reconfiguration network scenario where a feasible computational time is required.

In terms of the minimal loss achieved, HATSGA algorithm got results that are among the best ones found in the literature (second best) and, as such, does places this solution as an effective approach to compute power loss and find reconfiguration topologies. The elitism heuristics used within HATSGA proven to have a double positive aspect. Firstly, it minimizes the computational time required to compute the minimum loos topology that is essential for large scale and more complex topologies. Secondly, it drives the scalability of the computation based on real and effective electrical parameters configured in the algorithm itself. In other words, the ``expertise'' of the engineers is, fundamentally, the driver for the proposed heuristic.

The scalability of HATSGA for very large power distribution networks, although suggested as possible in this simulation run for IEEE 14-bus, is an aspect not fully addressed at this stage of the research and, as such, next step will include simulations using more complex and large-scale topologies.

\bibliographystyle{vancouver}
\bibliography{biblio}


\end{document}